\def\year{2021}\relax
\documentclass[letterpaper]{article} 
\usepackage{aaai21}  
\usepackage{times}  
\usepackage{helvet} 
\usepackage{courier}  
\usepackage[hyphens]{url}  
\usepackage{graphicx} 
\usepackage{natbib}  
\usepackage{caption} 
\usepackage[dvipsnames, svgnames, x11names]{xcolor}
\usepackage{amsmath}
\usepackage{amsfonts,amssymb}
\usepackage{multirow}
\usepackage{pifont}
\title{Bidirectional Machine Reading Comprehension for\\ Aspect Sentiment Triplet Extraction}
\author{
    Shaowei Chen$^1$, Yu Wang$^1$, Jie Liu$^{1,2}$\thanks{Corresponding author.}, Yuelin Wang$^1$\\

}
\affiliations{
    $^1$College of Artificial Intelligence, Nankai University, Tianjin, China\\
    $^2$Cloopen Research, Beijing, China\\
    \{shaoweichen, yuwang17, 1711496\}@mail.nankai.edu.cn \\
    jliu@nankai.edu.cn \\

}

\begin{document}

\maketitle

\begin{abstract}
Aspect sentiment triplet extraction (ASTE), which aims to identify aspects from review sentences along with their corresponding opinion expressions and sentiments, is an emerging task in fine-grained opinion mining. 
Since ASTE consists of multiple subtasks, including opinion entity extraction, relation detection, and sentiment classification, it is critical and challenging to appropriately capture and utilize the associations among them.
In this paper, we transform ASTE task into a multi-turn machine reading comprehension (MTMRC) task and propose a bidirectional MRC (BMRC) framework to address this challenge.
Specifically, we devise three types of queries, including \textit{non-restrictive extraction} queries, \textit{restrictive extraction} queries and \textit{sentiment classification} queries, to build the associations among different subtasks.
Furthermore, considering that an aspect sentiment triplet can derive from either an aspect or an opinion expression, we design a bidirectional MRC structure.
One direction sequentially recognizes aspects, opinion expressions, and sentiments to obtain triplets, while the other direction identifies opinion expressions first, then aspects, and at last sentiments. By making the two directions complement each other, our framework can identify triplets more comprehensively. 
To verify the effectiveness of our approach, we conduct extensive experiments on four benchmark datasets. The experimental results demonstrate that BMRC achieves state-of-the-art performances.

\end{abstract}
\section{Introduction}
\begin{figure}
\centering
\includegraphics[width=0.44\textwidth]{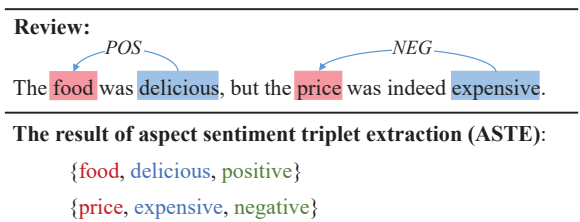}
\caption{An example of ASTE task. The aspects, opinion expressions, and sentiments are marked with red, blue, and green, respectively.}\label{intro}
\end{figure}

Fine-grained opinion mining is an important field in natural language processing (NLP). It comprises various tasks, such as aspect term extraction (ATE) \cite{DBLP:conf/emnlp/LiuXZ12,DECNN,Li2018,MaLWXW19}, opinion term extraction (OTE) \cite{DBLP:conf/naacl/FanWDHC19,Wu2020}, and aspect-level sentiment classification (ASC) \cite{DBLP:conf/ijcai/MaLZW17,DBLP:conf/emnlp/SunZMML19}. 
Existing studies generally solve these tasks individually or couple two of them as aspect and opinion terms co-extraction task \cite{DBLP:journals/tkde/LiuXZ15,Wang2016,DBLP:conf/aaai/WangPDX17,DBLP:conf/acl/DaiS19}, aspect term-polarity co-extraction task \cite{LuoLLZ19,LiBLL19}, and aspect-opinion pair extraction task \cite{SDRN,SpanMIT}. However, none of these studies can identify aspects, opinion expressions, and sentiments in a complete solution. 
To deal with this problem, the latest literature \cite{DBLP:conf/aaai/PengXBHLS20} presents aspect sentiment triplet extraction (ASTE) task, which aims to identify triplets such as (\textit{food, delicious, positive}) in Figure \ref{intro}.

Although these studies have achieved great progress, there are still several challenges existing in fine-grained opinion mining.
\textbf{First}, aspects and opinion expressions generally appear together in a review sentence and have explicit corresponding relations. Hence, how to adequately learn the association between ATE and OTE and make them mutually beneficial is a challenge. 
\textbf{Second}, the corresponding relations between aspects and opinion expressions can be complicated, such as one-to-many, many-to-one, and even overlapped and embedded. Thus, it is challenging to flexibly and exactly detect these relations. 
\textbf{Third}, each review sentence may contain multiple sentiments. For example, given the review in Figure \ref{intro}, the sentiments of \textit{price} and \textit{food} are negative and positive, respectively.
These sentiments are generally guided by the corresponding relations between aspects and opinion expressions.
Thus, how to properly introduce these relations to sentiment classification task is another challenge.

To address the aforementioned challenges, we deal with ASTE task and formalize it as a machine reading comprehension (MRC) task.
Given a query and a context, MRC task aims to capture the interaction between them and extract specific information from the context as the answer. 
Different from the general MRC task, we further devise multi-turn queries to identify aspect sentiment triplets due to the complexity of ASTE. Specially, we define this formalization as multi-turn machine reading comprehension (MTMRC) task.
By introducing the answers to the previous turns into the current turn as prior knowledge, the associations among different subtasks can be effectively learned.
For example, given the review in Figure \ref{intro}, we can identify the aspect \textit{food} in the first turn and introduce it into the second turn query \textit{What opinions given the aspect food?} to jointly identify the opinion expression \textit{delicious} and the relation between \textit{food} and \textit{delicious}. Then, we can use the aspect \textit{food} and the opinion expression \textit{delicious} as the prior knowledge of the third turn query to predict that the sentiment of \textit{food} is \textit{positive}. According to these turns, we can flexibly capture the association between ATE and OTE, detect complex relations between opinion entities\footnote{In this paper, we briefly note aspects and opinion expressions as opinion entities.}, and utilize these relations to guide sentiment classification.

Based on MTMRC, we propose a bidirectional machine reading comprehension (BMRC) framework\footnote{https://github.com/NKU-IIPLab/BMRC.} in this paper. 
Specifically, we design three-turn queries to identify aspect sentiment triplets. 
In the first turn, we design \textit{non-restrictive extraction} queries to locate the first entity of each aspect-opinion pair. 
Then, \textit{restrictive extraction} queries are designed for the second turn to recognize the other entity of each pair based on the previously extracted entity.
In the third turn, \textit{sentiment classification} queries are proposed to predict aspect-oriented sentiments based on the extracted aspects and their corresponding opinion expressions.
Since there is no intrinsic order when extracting aspects and opinion expressions, we further propose a bidirectional structure to recognize the aspect-opinion pairs. In one direction, we first utilize a non-restrictive extraction query to identify aspects such as $\left\{food, price\right\}$ in Figure \ref{model}. Then, given the specific aspect like \textit{food}, the second-turn query looks for its corresponding opinion expressions such as $\left\{delicious\right\}$ in Figure \ref{model} via a restrictive extraction query. Similarly, the other direction extracts opinion expressions and their corresponding aspects in a reversed order.
To verify the effectiveness of BMRC, we make comprehensive analyses on four benchmark datasets. The experimental results show that our approach substantially outperforms the existing methods. In summary, our contributions are three-fold:
\begin{itemize}
\item We formalize aspect sentiment triplet extraction (ASTE) task as a multi-turn machine reading comprehension (MTMRC) task. Based on this formalization, we can gracefully identify aspect sentiment triplets in a unified framework.
\item We propose a bidirectional machine reading comprehension (BMRC) framework. By devising three-turn queries, our model can effectively build the associations among opinion entity extraction, relation detection, and sentiment classification.
\item We conduct extensive experiments on four benchmark datasets. The experimental results demonstrate that our model achieves state-of-the-art performances.
\end{itemize}

\section{Related Work}
In this paper, we transform the aspect sentiment triplet extraction task into a multi-turn machine reading comprehension task. Thus, we introduce the related work from two parts, including fine-grained opinion mining and machine reading comprehension.
\subsection{Fine-grained Opinion Mining} 

Fine-grained opinion mining consists of various tasks, including aspect term extraction (ATE) \cite{Wang2016, DBLP:conf/acl/HeLND17, Li2018, DECNN, DBLP:conf/emnlp/LiL17}, opinion term extraction (OTE) \cite{liu2015fine, poria2016aspect, DECNN, Wu2020}, aspect-level sentiment classiﬁcation (ASC) \cite{DBLP:conf/acl/DongWTTZX14, DBLP:conf/emnlp/TangQL16, DBLP:conf/aaai/LiW0Z019, DBLP:conf/acl/HeLND18, DBLP:conf/naacl/HazarikaPVKCZ18, DBLP:conf/emnlp/NguyenS15, DBLP:conf/acl/LiuCWMZ18}, etc. The studies solve these tasks individually and ignore the dependency between them.

To explore the interactions between different tasks, recent studies gradually focus on the joint tasks such as aspect term-polarity co-extraction \cite{he2019interactive, DBLP:conf/emnlp/MitchellAWD13, DBLP:conf/aaai/LiL17, LiBLL19}, aspect and opinion terms co-extraction \cite{DBLP:journals/tkde/LiuXZ15,Wang2016, DBLP:conf/aaai/WangPDX17,DBLP:conf/acl/DaiS19}, aspect category and sentiment classification \cite{DBLP:conf/emnlp/HuZZCSCS19}, and aspect-opinion pair extraction \cite{SDRN,SpanMIT}. Besides, there are also a lot of studies \cite{RACL2020,he2019interactive} solving multiple tasks with a multi-task learning network. However, none of these studies could identify aspects, opinion expressions and sentiments in a unified framework.
To deal with this issue, Peng et al. \shortcite{DBLP:conf/aaai/PengXBHLS20} proposed a two-stage framework to solve aspect sentiment triplet extraction (ASTE) task, which aims to extract triplets of aspects, opinion expressions and sentiments. However, the model suffers from error propagation due to its two-stage framework. Besides, separating the extraction and pairing of opinion entities means that the associations between different tasks are still not adequately considered.

\subsection{Machine Reading Comprehension}
Machine reading comprehension (MRC) aims to answer specific queries based on a given context. Recent researches have proposed various effective architectures for MRC, which adequately learn the interaction between the query and context. For example, BiDAF \cite{DBLP:conf/iclr/SeoKFH17} employs a RNN-based sequential framework to encode queries and passages, while QANet \cite{DBLP:conf/iclr/YuDLZ00L18} employs both convolution and self-attention. Several MRC systems \cite{DBLP:conf/naacl/PetersNIGCLZ18, radford2019language} adopt context-aware embedding as well and obtain comparable results, especially BERT-based MRC model \cite{DBLP:conf/naacl/DevlinCLT19}. 

Recently, there is a tendency to apply MRC on many NLP tasks, including named entity recognition \cite{li2019unified}, entity relation extraction \cite{DBLP:conf/acl/LiYSLYCZL19,DBLP:conf/conll/LevySCZ17}, and summarization \cite{DBLP:journals/corr/abs-1806-08730}, etc.
Due to the advantages of MRC framework, we naturally transform ASTE into a multi-turn MRC task to better construct the associations among aspects, opinions, aspect-opinion relations and sentiments through well-designed queries.
Different from the existing methods \cite{li2019unified,DBLP:conf/acl/LiYSLYCZL19}, we innovatively propose a bidirectional framework, which can identity triplets more comprehensively by making the two directions complement each other. This framework can be further extended to other tasks such as entity relation extraction.


\begin{figure}
\centering
\includegraphics[width=0.45\textwidth]{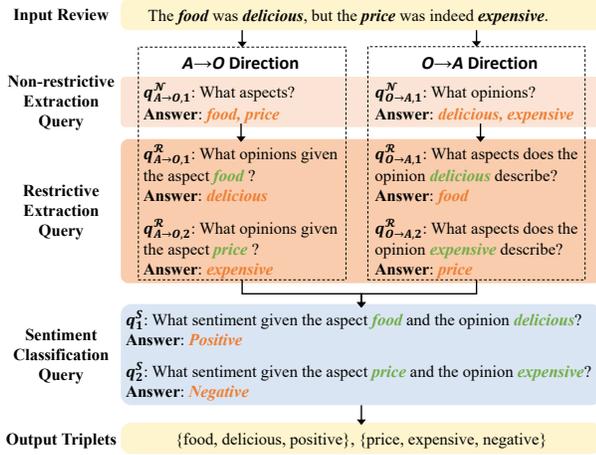}
\caption{The bidirectional machine reading comprehension (BMRC) framework.}\label{model}
\end{figure}

\section{Problem Formulation}
Given a review sentence $X=\left\{x_{1}, x_{2},..., x_{N}\right\}$ with $N$ tokens, ASTE task aims to identify the collection of triplets $T=\left\{\left(a_{i}, o_{i}, s_{i}\right)\right\}_{i=1}^{\left |  T\right |}$, where $a_{i}$, $o_{i}$, $s_{i}$, and $\left |  T\right |$ represent the aspect, the opinion expression, the sentiment, and the number of triplets\footnote{The $\left |*\right |$ represents the number of elements in the collection $*$.}, respectively.

To formalize ASTE task as a multi-turn MRC task, we construct three types of queries\footnote{We use superscripts $\mathcal{N}$, $\mathcal{R}$, and $\mathcal{S}$ to denote the query types.}, including non-restrictive extraction queries $Q^{\mathcal{N}}=\left\{q_{i}^{\mathcal{N}}\right\}_{i=1}^{\left |  Q^{\mathcal{N}}\right |}$, restrictive extraction queries $Q^{\mathcal{R}}=\left\{q_{i}^{\mathcal{R}}\right\}_{i=1}^{\left |  Q^{\mathcal{R}}\right |}$ and sentiment classification queries $Q^{\mathcal{S}}=\left\{q_{i}^{\mathcal{S}}\right\}_{i=1}^{\left |  Q^{\mathcal{S}}\right |}$. 

Concretely, in the first turn, each non-restrictive extraction query $q_{i}^{\mathcal{N}}$ aims to extract either aspects $A=\left\{a_{i}\right\}_{i=1}^{\left |  A\right |}$ or opinion expressions $O=\left\{o_{i}\right\}_{i=1}^{\left |  O\right |}$ from the review sentence to trigger aspect-opinion pairs.
In the second turn, given the opinion entities recognized by $q_{i}^{\mathcal{N}}$, each restrictive extraction query $q_{i}^{\mathcal{R}}$ aims to identify either the corresponding aspects or the corresponding opinion expressions. To be more specific, given each aspect $a_{i}$ extracted by $q_{i}^{\mathcal{N}}$, the restrictive extraction query extracts its corresponding opinion expressions $O_{a_{i}}=\left\{o_{a_{i},j}\right\}_{j=1}^{\left |  O_{a_{i}}\right |}$.
In the final turn, each sentiment classification query $q_{i}^{\mathcal{S}}$ predicts the sentiment $s_{a_{i}}\in \left\{\text{Positive}, \text{Negative}, \text{Neutral}\right\}$ for each aspect $a_{i}$.

\section{Methodology}
\subsection{Framework}
To deal with ASTE task, we propose a bidirectional machine reading comprehension (BMRC) framework. The overall framework is illustrated in Figure \ref{model}. Concretely, we first design non-restrictive extraction queries and restrictive extraction queries to extract aspect-opinion pairs. Considering that each pair can be triggered by an aspect or an opinion expression, we further construct a bidirectional structure.
In one direction, the aspects are first extracted via non-restrictive extraction queries, and then the corresponding opinion expressions for each aspect are identified via restrictive extraction queries. We define the above process as A$\rightarrow$O direction. Similarly, in O$\rightarrow$A direction, the framework recognizes opinion expressions and their corresponding aspects in a reversed order.
After that, we design sentiment classification queries to predict the sentiment polarity for each aspect. Furthermore, our model jointly learns to answer the above queries to make them mutually beneficial. 
Besides, we adopt BERT as the encoding layer for richer semantics representations. During inference, the model fuses the answers to different queries and forms the triplets.

\subsection{Query Construction}
In BMRC, we adopt a template-based approach to construct queries.
Specifically, we first design the non-restrictive extraction query and the restrictive extraction query in the A$\rightarrow$O direction as follows:    
\begin{itemize}
	\item \textbf{A$\rightarrow$O non-restrictive extraction query} $q_{A\rightarrow O}^{\mathcal{N}}$: We design query `\textit{What aspects?}' to extract the collection of aspects $A=\left\{a_{i}\right\}_{i=1}^{\left |  A\right |}$ from the given review sentence $X$.
	\item \textbf{A$\rightarrow$O restrictive extraction query} $q_{A\rightarrow O}^{\mathcal{R}}$: We design query `\textit{What opinions given the aspect $a_{i}$?}' to extract the corresponding opinions $O_{a_{i}}=\left\{o_{a_{i},j}\right\}_{j=1}^{\left |  O_{a_{i}}\right |}$ for each aspect $a_{i}$ and form aspect-opinion pairs.
\end{itemize}

Reversely, the O$\rightarrow$A direction extraction queries are constructed as follows:
\begin{itemize}
	
	\item \textbf{O$\rightarrow$A non-restrictive extraction query} $q_{O\rightarrow A}^{\mathcal{N}}$: We use query `\textit{What opinions?}' to extract the collection of opinion expressions $O=\left\{o_{i}\right\}_{i=1}^{\left |  O\right |}$.

	\item \textbf{O$\rightarrow$A restrictive extraction query} $q_{O\rightarrow A}^{\mathcal{R}}$: We design query `\textit{What aspect does the opinion $o_{i}$ describe?}' to recognize the corresponding aspects $A_{o_{i}}=\left\{a_{o_{i},j}\right\}_{j=1}^{\left |  A_{o_{i}}\right |}$ for each opinion expression $o_{i}$.

\end{itemize}

With the above queries, opinion entity extraction and relation detection are naturally fused, and the dependency between them is gracefully learned via the restrictive extraction queries.
Then, we devise sentiment classification queries to classify the aspect-oriented sentiments as follows:
\begin{itemize}
	\item \textbf{Sentiment Classification query} $q^{\mathcal{S}}$: We design query `\textit{What sentiment given the aspect $a_{i}$ and the opinion $o_{a_{i},1}/.../o_{a_{i},\left |  O_{a_{i}}\right |}$?}' to predict sentiment polarity $s_{a_{i}}$ for each aspect $a_{i}$.
\end{itemize}

With sentiment classification queries, the semantics of aspects and their corresponding opinion expressions can be adequately considered during sentiment prediction.

\subsection{Encoding Layer}
Given the review sentence $X=\left\{x_{1}, x_{2},..., x_{N}\right\}$ with $N$ tokens and each query $q_{i}=\{q_{i,1},q_{i,2},...,q_{i,\left|q_{i}\right|}\}$ with $\left|q_{i}\right|$ tokens, the encoding layer learns the context representation for each token. Inspired by the successful practice on many NLP tasks, we adopt BERT as the encoder. 
Formally, we first concatenate the query $q_{i}$ and the review sentence $X$ to obtain the combined input $I=\{[{\rm CLS}],q_{i,1},q_{i,2},...,q_{i,\left|q_{i}\right|},[{\rm SEP}],x_{1}, x_{2},..., x_{N}\}$, where $[{\rm CLS}]$ and $[{\rm SEP}]$ are the beginning token and the segment token. The initial representation $\mathbf{e}_{i}$ for each token is constructed by summing its word embedding $\mathbf{e}_{i}^{w}$, position embedding $\mathbf{e}_{i}^{p}$, and segment embedding $\mathbf{e}_{i}^{g}$. Then, BERT is used to encode the initial representation sequence $E=\left\{\mathbf{e}_{1}, \mathbf{e}_{2},...,\mathbf{e}_{\left|q_{i}\right|+N+2}\right\}$ as the hidden representation sequence $H=\left\{\mathbf{h}_{1}, \mathbf{h}_{2},...,\mathbf{h}_{\left|q_{i}\right|+N+2}\right\}$ with the stacked Transformer blocks. 

\subsection{Answer Prediction}
\subsubsection{Answer for Extraction Query}
For non-restrictive and restrictive extraction queries, the answers could be multiple opinion entities extracted from the review sentence $X$. For example, given the review in Figure \ref{model}, the aspects \textit{food} and \textit{price} should be extracted as the answer to the A$\rightarrow$O non-restrictive extraction query $q_{A\rightarrow O, 1}^{\mathcal{N}}$. Thus, we utilize two binary classifiers to predict the answer spans. Specifically, based on the hidden representation sequence $H$, one classifier predicts whether each token $x_{i}$ is the start position of the answer or not, and another predicts the possibility that each token is the end position:
\begin{equation}
    p\left(y_{i}^{start}|x_{i}, q\right) = {\rm softmax}(\mathbf{h}_{\left|q\right|+2+i}W_{s}),
\end{equation}
\begin{equation}
    p\left(y_{i}^{end}|x_{i}, q\right) = {\rm softmax}(\mathbf{h}_{\left|q\right|+2+i}W_{e}),
\end{equation}
where $W_{s}\in \mathbf{R}^{d_{h}\times 2}$ and $W_{e}\in \mathbf{R}^{d_{h}\times 2}$ are model parameters, $d_{h}$ denotes the dimension of hidden representations in $H$, and $\left|q\right|$ is the query length.

\subsubsection{Answer for Sentiment Classification Query}
Following existing work \cite{DBLP:conf/naacl/DevlinCLT19}, the answer to sentiment classification query is predicted with the hidden representation of $\left[{\rm CLS}\right]$. Formally, we append a three-class classifier to BERT for predicting the sentiment $y^{\mathcal{S}}$ as follows:
\begin{equation}
    p\left(y^{\mathcal{S}}|X, q\right) = {\rm softmax}(\mathbf{h}_{1}W_{c}),
\end{equation}
where $W_{c}\in \mathbf{R}^{d_{h}\times 3}$ is model parameter.

\subsection{Joint Learning}
To jointly learn the subtasks in ASTE and make them mutually beneficial, we fuse the loss functions of different queries. For non-restrictive extraction queries in both two directions, we minimize the cross-entropy loss as follows:
\begin{small}
\begin{equation}
\begin{aligned}
    \mathcal{L}_{\mathcal{N}}=-\sum_{i=1}^{\left |  Q^{\mathcal{N}}\right |}\sum_{j=1}^{N}[p\left(y_{j}^{start}|x_{j}, q_{i}^{\mathcal{N}}\right){\rm log}\hat{p}\left(y_{j}^{start}|x_{j}, q_{i}^{\mathcal{N}}\right)\\+p\left(y_{j}^{end}|x_{j}, q_{i}^{\mathcal{N}}\right){\rm log}\hat{p}\left(y_{j}^{end}|x_{j}, q_{i}^{\mathcal{N}}\right)],
\end{aligned}
\end{equation}
\end{small}
where $p\left(*\right)$ represents the gold distribution, and $\hat{p}\left(*\right)$ denotes the predicted distribution. 

Similarly, the loss of restrictive extraction queries in both two directions is calculated as follows:
\begin{small}
\begin{equation}
\begin{aligned}
    \mathcal{L}_{\mathcal{R}}=-\sum_{i=1}^{\left |  Q^{\mathcal{R}}\right |}\sum_{j=1}^{N}[p\left(y_{j}^{start}|x_{j}, q_{i}^{\mathcal{R}}\right){\rm log}\hat{p}\left(y_{j}^{start}|x_{j}, q_{i}^{\mathcal{R}}\right)\\+p\left(y_{j}^{end}|x_{j}, q_{i}^{\mathcal{R}}\right){\rm log}\hat{p}\left(y_{j}^{end}|x_{j}, q_{i}^{\mathcal{R}}\right)].
\end{aligned}
\end{equation}
\end{small}

For the sentiment classification queries, we minimize the cross-entropy loss function as follows:
\begin{small}
\begin{equation}
    \mathcal{L}_{\mathcal{S}} = -\sum_{i=1}^{\left |  Q^{\mathcal{S}}\right |}p\left(y^{\mathcal{S}}|X,q_{i}^{\mathcal{S}}\right){\rm log}\hat{p}\left(y^{\mathcal{S}}|X,q_{i}^{\mathcal{S}}\right).
\end{equation}
\end{small}

Then, we combine the above loss functions to form the loss objective of the entire model:
\begin{equation}
    \mathcal{L}\left(\theta\right) = \mathcal{L}_{\mathcal{N}}+\mathcal{L}_{\mathcal{R}}+\mathcal{L}_{\mathcal{S}}.
\end{equation}

The optimization problem in Eq.(7) can be solved by any gradient descent approach. In this paper, we adopt the AdamW \cite{DBLP:journals/corr/abs-1711-05101} approach.

\subsection{Inference}
During inference, we fuse the answers to different queries to obtain triplets. Specifically, in the A$\rightarrow$O direction, the non-restrictive extraction query $q_{A \rightarrow O}^{\mathcal{N}}$ first identifies the aspect collection $A=\left\{a_{1}, a_{2},..., a_{\left |  A\right |}\right\}$ with $\left |  A\right |$ aspects. For each predicted aspect $a_{i}$, the A$\rightarrow$O restrictive query $q_{A\rightarrow O, i}^{\mathcal{R}}$ recognizes the corresponding opinion expression collection and obtains the set of predicted aspect-opinion pairs $V_{A\rightarrow O}=\left[\left(a_{k}, o_{k}\right)\right]_{k=1}^{K}$ in the A$\rightarrow$O direction. Similarly, in the O$\rightarrow$A direction, the model identifies the set of aspect-opinion pairs $V_{O\rightarrow A}=\left[\left(a_{l}, o_{l}\right)\right]_{l=1}^{L}$ in a reversed order.
Then, we combine $V_{A\rightarrow O}$ and $V_{O\rightarrow A}$ as follows:
\begin{equation}
	V ={V}'\cup \left\{\left(a,o\right)|\left(a,o\right)\in {V}'',p\left(a,o\right)>\delta \right\},
\end{equation}
\begin{equation}
	p\left(a,o\right) =\left\{\begin{matrix}
 p\left ( a \right )p\left ( o|a \right )& if \left ( a,o \right )\in V_{A\rightarrow O}\\ 
 p\left ( o \right )p\left ( a|o \right )& if \left ( a,o \right )\in V_{O\rightarrow A}
\end{matrix}\right.,
\end{equation}
where ${V}'$ and ${V}''$ denote the intersection and difference set of $V_{A\rightarrow O}$ and $V_{O\rightarrow A}$, respectively. Each aspect-opinion pair in ${V}''$ is valid only if its probability $p\left(a,o\right)$ is higher than the given threshold $\delta$. The probability of each opinion entity is calculated by multiplying the probabilities of its start and end positions.

Finally, we construct sentiment classification query $q_{i}^{\mathcal{S}}$ to predict the sentiment $s_{a_{i}}$ of each aspect $a_{i}$. Based on these, the triplet collection $T=\left[\left(a_{i}, o_{i}, s_{i}\right)\right]_{i=1}^{\left |  T\right |}$ can be obtained.

\begin{table}[]
\centering
\scalebox{0.8}{
\begin{tabular}{c|c|c|c|c|c|c}
\hline
\multicolumn{1}{c|}{\multirow{2}{*}{Datasets}} & \multicolumn{2}{c|}{Train}& \multicolumn{2}{c|}{Dev}& \multicolumn{2}{c}{Test}\\ \cline{2-7} 
&\#S & \#T &\#S & \#T&\#S & \#T\\ \hline
14-Lap \cite{DBLP:conf/semeval/PontikiGPPAM14}	& 920 &1265 & 228 &337  & 339 &490\\
14-Res \cite{DBLP:conf/semeval/PontikiGPPAM14}  & 1300& 2145& 323 & 524 & 496 & 862\\
15-Res \cite{DBLP:conf/semeval/PontikiGPMA15}  & 593 & 923 & 148 & 238 & 318 & 455\\
16-Res \cite{DBLP:conf/semeval/PontikiGPAMAAZQ16}  & 842 & 1289& 210 & 316 & 320 & 465 \\ \hline
\end{tabular}}
\caption{Statistics of datasets. \#S and \#T denote the number of sentences and triplets, respectively. }
\label{Statistic}
\end{table}
\section{Experiments}

\begin{table*}[]
\centering
\scalebox{0.75}{
\begin{tabular}{l|l|cccc|cccc|cccc|cccc}
\hline
\multicolumn{1}{c|}{\multirow{2}{*}{Evaluation}} & \multicolumn{1}{c|}{\multirow{2}{*}{Models}} & \multicolumn{4}{c|}{14-Lap} & \multicolumn{4}{c|}{14-Res} & \multicolumn{4}{c|}{15-Res} & \multicolumn{4}{c}{16-Res} \\ \cline{3-18}
 &       & A-S       & O       & P      & T       & A-S       & O       & P      & T      & A-S       & O       & P      & T      & A-S       & O       & P      & T     \\ \hline
 \multicolumn{1}{c|}{\multirow{5}{*}{Precision}}  &TSF         &  63.15   & 78.22  & 50.00   & 40.40   & 76.60   & 84.72   &  47.76   & 44.18    & 67.65    &  78.07   & 49.22      & 40.97 & 71.18   &  81.09   & 52.35      & 46.76  \\ 
&RINANRTE+       & 41.20 & 78.20 & 34.40 & 23.10 & 48.97 & 81.06 & 42.32 & 31.07 & 46.20 & 77.40 & 37.10 & 29.40 & 49.40 & 75.00 & 35.70 & 27.10   \\ 
&Li-unified-R+   & 66.28 & 76.62 & 52.29 & 42.25 & 73.15 & 81.20 & 44.37 & 41.44 & 64.95 & 79.18 & 52.75 & 43.34 & 66.33 & 79.84 & 46.11 & 38.19
  \\ 
&RACL+R     &   59.75   & 77.58   &  54.22 & 41.99   &  75.57   &  82.28   &  73.58  &  62.64   &  68.35  &  76.25  &  67.89   & 55.45  &   68.53  &   82.52  &  72.77     & 60.78  \\
&Ours &\textbf{72.73}    &  \textbf{84.67}    &  \textbf{74.11}    &  \textbf{65.12}     &  \textbf{77.74}    &  \textbf{87.22}    &  \textbf{76.91}    &  \textbf{71.32}    &  \textbf{72.41}    &  \textbf{82.99}    &  \textbf{71.59}    &  \textbf{63.71}    &  \textbf{73.69}     &  \textbf{85.31}    &  \textbf{76.08}    &  \textbf{67.74}  
 \\ \hline
 \multicolumn{1}{c|}{\multirow{5}{*}{Recall}}  & TSF         & 61.55 & 71.84 & 58.47 & 47.24 & 67.84 & 80.39 & 68.10 & 62.99 & 64.02 &  78.07    & 65.70 & 54.68 & 72.30 &  86.67    & 70.50 & 62.97 \\ 
&RINANRTE+             & 33.20 & 62.70 & 26.20 & 17.60 & 47.36 & 72.05 & 51.08 & 37.63 & 37.40 & 57.00 & 33.90 & 26.90 & 36.70 & 42.40 & 27.00 & 20.50  \\ 
&Li-unified-R+     & 60.71 &  74.90     & 52.94 & 42.78 & 74.44 & 83.18 & 73.67 & 68.79 &  64.95   & 75.88 & 61.75 & 50.73 &  74.55    & 86.88 & 64.55 & 53.47 \\ 
&RACL+R     &   \textbf{68.90}   & \textbf{81.22}   & \textbf{66.94}  &  51.84  &  \textbf{82.23}  &  \textbf{90.49}   &  67.87  &  57.77   &   \textbf{70.72}  & \textbf{83.96}   &  63.74   & 52.53  &  \textbf{78.52}   &   \textbf{91.40}  &   71.83    & 60.00  \\
&Ours   &  62.59    & 67.18 &  61.92    &  \textbf{54.41}    &  75.10    &  82.90    &  \textbf{75.59}    &  \textbf{70.09}  & 62.63 & 73.23 &  \textbf{65.89}    &  \textbf{58.63}   & 72.69 & 83.01 &  \textbf{76.99}    &  \textbf{68.56}   \\ \hline
\multicolumn{1}{c|}{\multirow{5}{*}{F$_{1}$-score}}  &TSF         & 62.34 & 74.84 & 53.85 & 43.50 & 71.95 & 82.45 & 56.10 & 51.89 & 65.79 &  78.02    & 56.23 & 46.79 & 71.73 & 83.73 & 60.04 & 53.62\\ 
&RINANRTE+             & 36.70 & 69.60 & 29.70 & 20.00 & 48.15 & 76.29 & 46.29 & 34.03 & 41.30 & 65.70 & 35.40 & 28.00 & 42.10 & 54.10 & 30.70 & 23.30  \\ 
&Li-unified-R+     & 63.38 & 75.70 & 52.56 & 42.47 & 73.79 & 82.13 & 55.34 & 51.68 & 64.95 & 77.44 & 56.85 & 46.69 & 70.20 & 83.16 & 53.75 & 44.51 \\ 
&RACL+R     &   64.00   &  \textbf{79.36}  & 59.90  & 46.39   & \textbf{78.76}    &  \textbf{86.19}   & 70.61   & 60.11    & \textbf{69.51}   & \textbf{79.91}   & 65.46    & 53.95  &  \textbf{73.19}   & \textbf{86.73}    &   72.29    & 60.39  \\
&Ours   &   \textbf{67.27}   &  74.90   &  \textbf{67.45}   &  \textbf{59.27}   &  76.39    &  84.99    &  \textbf{76.23}    &  \textbf{70.69}    &  67.16    & 77.79 &  \textbf{68.60}    &  \textbf{61.05}    &  73.18   &  84.13  &  \textbf{76.52}  &  \textbf{68.13}   \\ \hline
\end{tabular}}
\caption{Experimental results (\%). Specifically,  `A-S', `O', `P', and `T' denote aspect term and sentiment co-extraction, opinion term extraction, aspect-opinion pair extraction, and aspect sentiment triplet extraction, respectively. }\label{precision}
\end{table*}

\subsection{Datasets}

To verify the effectiveness of our proposed approach, we conduct experiments on four benchmark datasets\footnote{https://github.com/xuuuluuu/SemEval-Triplet-data} from the SemEval ABSA Challenges \cite{DBLP:conf/semeval/PontikiGPPAM14, DBLP:conf/semeval/PontikiGPMA15, DBLP:conf/semeval/PontikiGPAMAAZQ16} and list the statistics of these datasets in Table \ref{Statistic}. Specifically, the golden annotations for opinion expressions and relations are derived from \citet{DBLP:conf/naacl/FanWDHC19}. And we split the datasets as \citet{DBLP:conf/aaai/PengXBHLS20} did.

\subsection{Experimental Settings}
For the encoding layer, we adopt the \textbf{BERT-base} \cite{DBLP:conf/naacl/DevlinCLT19} model with 12 attention heads, 12 hidden layers and the hidden size of 768, resulting into 110M pretrained parameters. During training, we use AdamW \cite{DBLP:journals/corr/abs-1711-05101} for optimization with weight decay 0.01 and warmup rate 0.1. The learning rate for training classifiers and the fine-tuning rate for BERT are set to 1e-3 and 1e-5 respectively. Meanwhile, we set batch size to 4 and dropout rate to 0.1. According to the triplet extraction F$_{1}$-score on the development sets, the threshold $\delta$ is manually tuned to 0.8 in bound $[0, 1)$ with step size set to 0.1. 
We run our model on a Tesla V100 GPU and train our model for 40 epochs in about 1.5h.

\subsection{Evaluation}
To comprehensively measure the performances of our model and the baselines, we use \textit{Precision}, \textit{Recall}, and \textit{F$_{1}$-score} to evaluate the results on four subtasks, including aspect term and sentiment co-extraction, opinion term extraction, aspect-opinion pair extraction, and triplet extraction.
For reproducibility, we report the testing results averaged over 5 runs with different random seeds. At each run, we select the testing results when the model achieves the best performance on the development set.

\subsection{Baselines}
To demonstrate the effectiveness of BMRC, we compare our model with the following baselines:
\begin{itemize}
    \item \textbf{TSF} \cite{DBLP:conf/aaai/PengXBHLS20} is a two-stage pipeline model for ASTE. In the first stage, TSF extracts both aspect-sentiment pairs and opinion expressions. In the second stage, TSF pairs up the extraction results into triplets via an relation classifier.
    \item \textbf{RINANTE+} adopts RINANTE \cite{DBLP:conf/acl/DaiS19} with additional sentiment tags as the first stage model to joint extract aspects, opinion expressions, and sentiments. Then, it adopts the second stage of TSF to detect the corresponding relations between opinion entities.
    \item \textbf{Li-unified-R+} jointly identifies aspects and their sentiments with Li-unified \cite{LiBLL19}. Meanwhile, it predicts opinion expressions with an opinion-enhanced component at the first stage. Then, it also uses the second stage of TSF to predicts relations.
    \item \textbf{RACL+R} first adopts RACL \cite{RACL2020} to identify the aspects, opinion expressions, and sentiments. Then, we construct the query 'Matched the aspect $a_{i}$ and the opinion expression $o_{j}$?' to detect the relations. Note that RACL is also based on BERT.


\end{itemize}


\subsection{Results}
The experimental results are shown in Table \ref{precision}.
According to the results, our model achieves state-of-the-art performances on all datasets. 
Although the improvements on aspect term and sentiment co-extraction and opinion term extraction are slight, our model significantly surpasses the baselines by an average of 5.14\% F$_{1}$-score on aspect-opinion pair extraction and an average of 9.58\% F$_{1}$-score on triplet extraction. 
The results indicate that extracting opinion entities and relations in pipeline will lead to severe error accumulation. By utilizing the BMRC framework, our model effectively fuses and simplifies the tasks of ATE, OTE, and relation detection, and avoids the above issue.
It is worth noting that the increase in precision contributes most to the boost of F1-score, which shows that the predictions of our model own higher reliability than those baselines. Besides, RACL+R outperforms than other baselines because BERT can learn richer context semantics. TSF and Li-unified-R+ achieve better performances than RINANTE+ because TSF and Li-unified-R+ introduce complex mechanisms to solve the issue of sentiment contradiction brought by the unified tagging schema. Different from those approaches, our model gracefully solves this issue by transforming ASTE into a multi-turn MRC task.

Considering that the datasets released by \citet{DBLP:conf/aaai/PengXBHLS20} remove the cases that one opinion expression corresponds to multiple aspects, we also conduct experiments on AFOE datasets \footnote{https://github.com/NJUNLP/GTS} \cite{Wu2020EMNLP} and report the results in Table \ref{AFOE}. The AFOE datasets, which retains the above cases, are also constructed based on the datasets of \citet{DBLP:conf/naacl/FanWDHC19} and the original SemEval ABSA Challenges. And we further compare our model with two baselines, including IMN+IOG and GTS\footnote{It worth noting that this paper has not been published when we submit our paper to AAAI 2021.} \cite{Wu2020EMNLP} . Specifically, IMN+IOG is a pipeline model which utilizes the interactive multi-task learning network (IMN) \cite{he2019interactive} as the first stage model to identity the aspects and their sentiments. Then, IMN+IOG use the Inward-Outward LSTM \cite{DBLP:conf/naacl/FanWDHC19} as the second stage model to extract the aspect-oriented opinion expressions. And GTS is a latest model which proposes a grid tagging schema to identify the aspect sentiment triplets in an end-to-end way. Particularly, GTS also utilizes BERT as the encoder and designs an inference strategy to exploit mutual indication between different opinion factors. According to the results, our model and GTS significantly outperform IMN+IOG because the joint methods can solve the error propagation problem. Compared with GTS, our model still achieves competitive performances, which verify the effectiveness of our model.

\section{Ablation Study}
To further validate the origination of the significant improvement of BMRC, we conduct ablation experiments and answer the following questions:
\begin{itemize}
	\item Does the restrictive extraction query build the association between opinion entity extraction and relation detection?
	\item Does the bidirectional structure promote the performance of aspect-opinion pair extraction?
    \item Do the relations between aspects and opinion expressions enhance the sentiment classification?
	\item How much improvement can the BERT bring? 
\end{itemize}



\subsection{Effect of the Restrictive Extraction Query}
\label{dependency}
\begin{figure*}
\centering
\begin{minipage}{0.35\textwidth}
\centering
\includegraphics[width=1\textwidth]{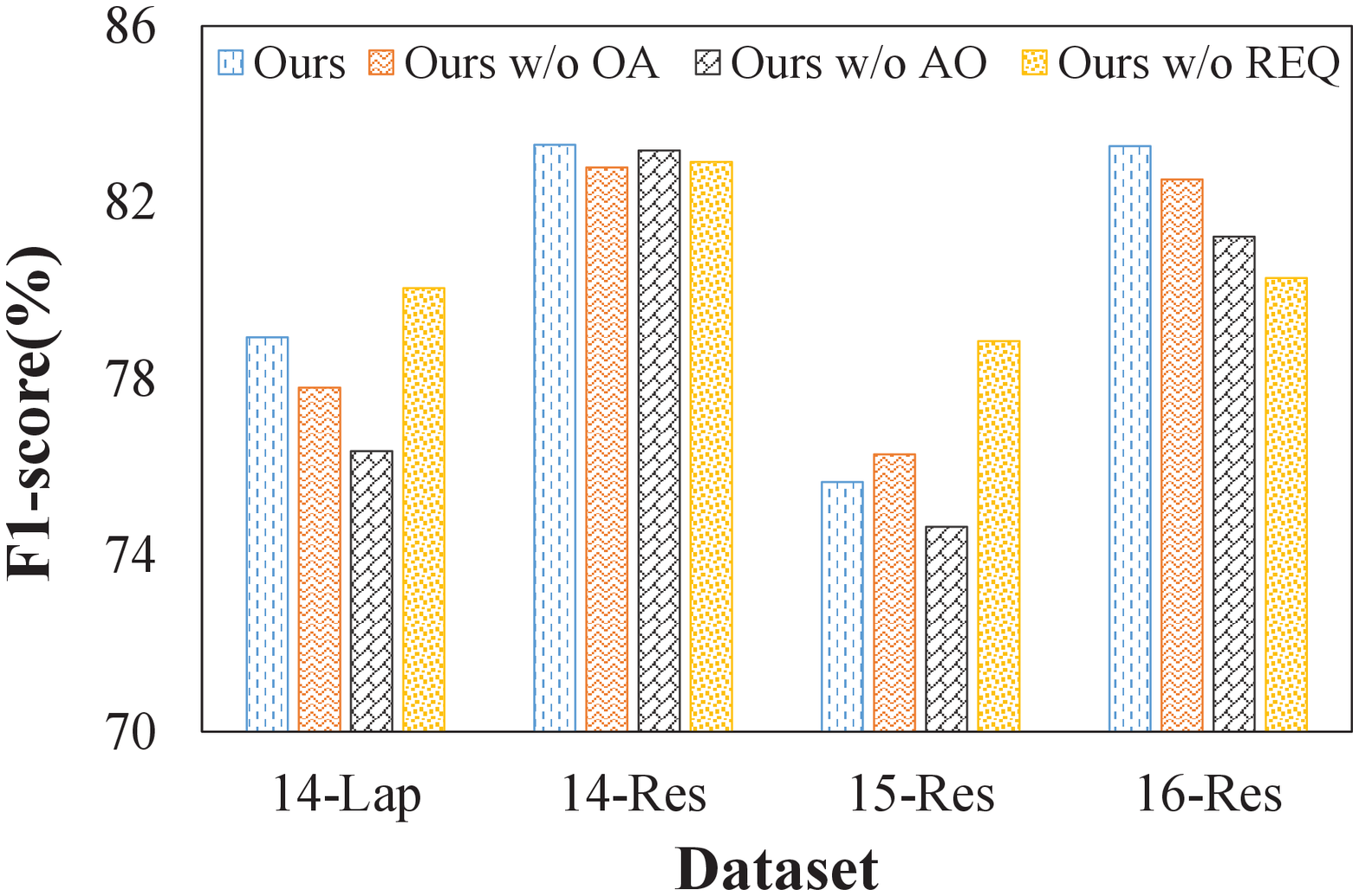}
(a) Aspect Term Extraction\\
\end{minipage}
\begin{minipage}{0.35\textwidth}
\centering
\includegraphics[width=1\textwidth]{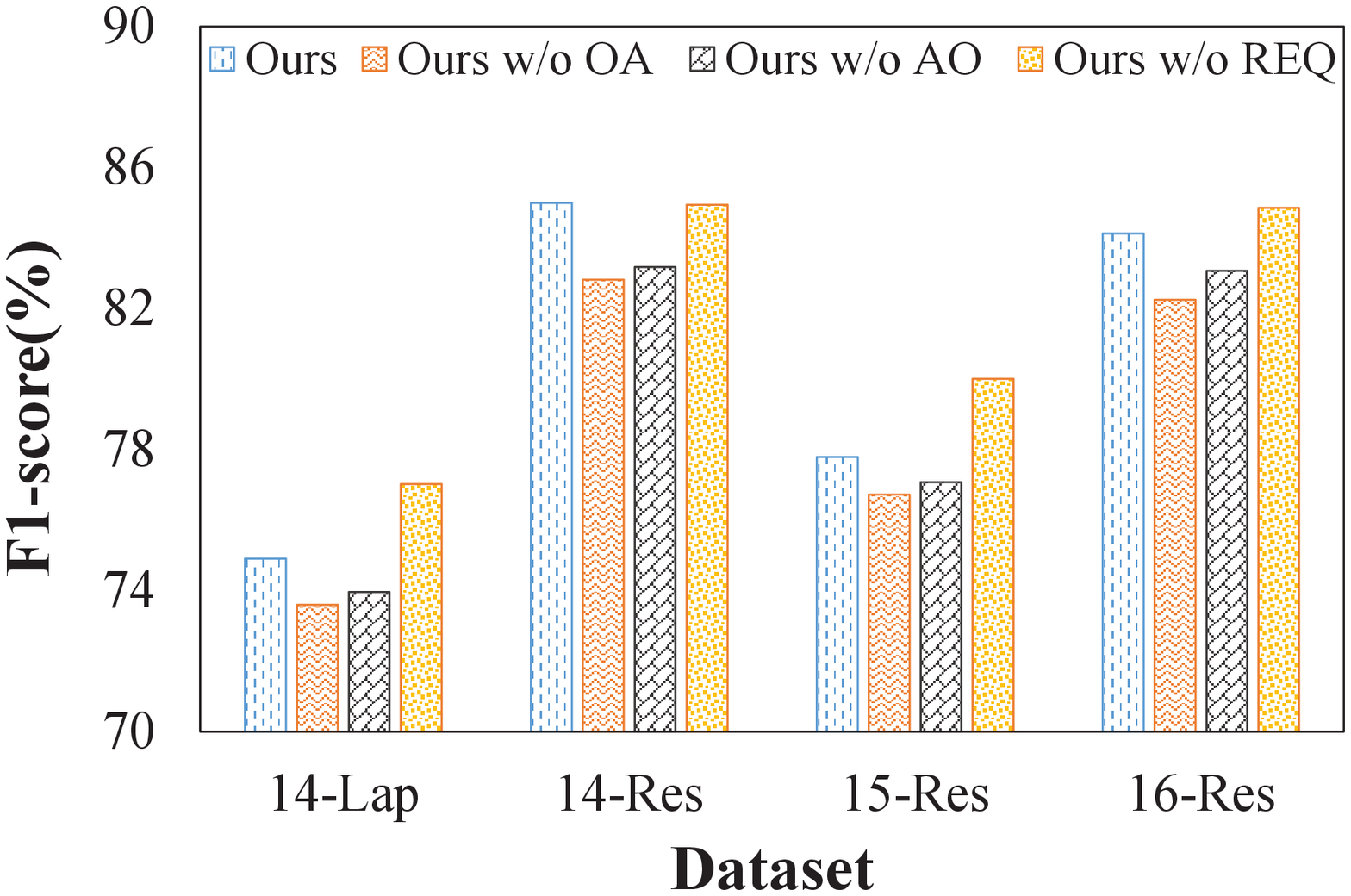}
(b) Opinion Term Extraction\\
\end{minipage}
\begin{minipage}{0.35\textwidth}
\centering
\includegraphics[width=1\textwidth]{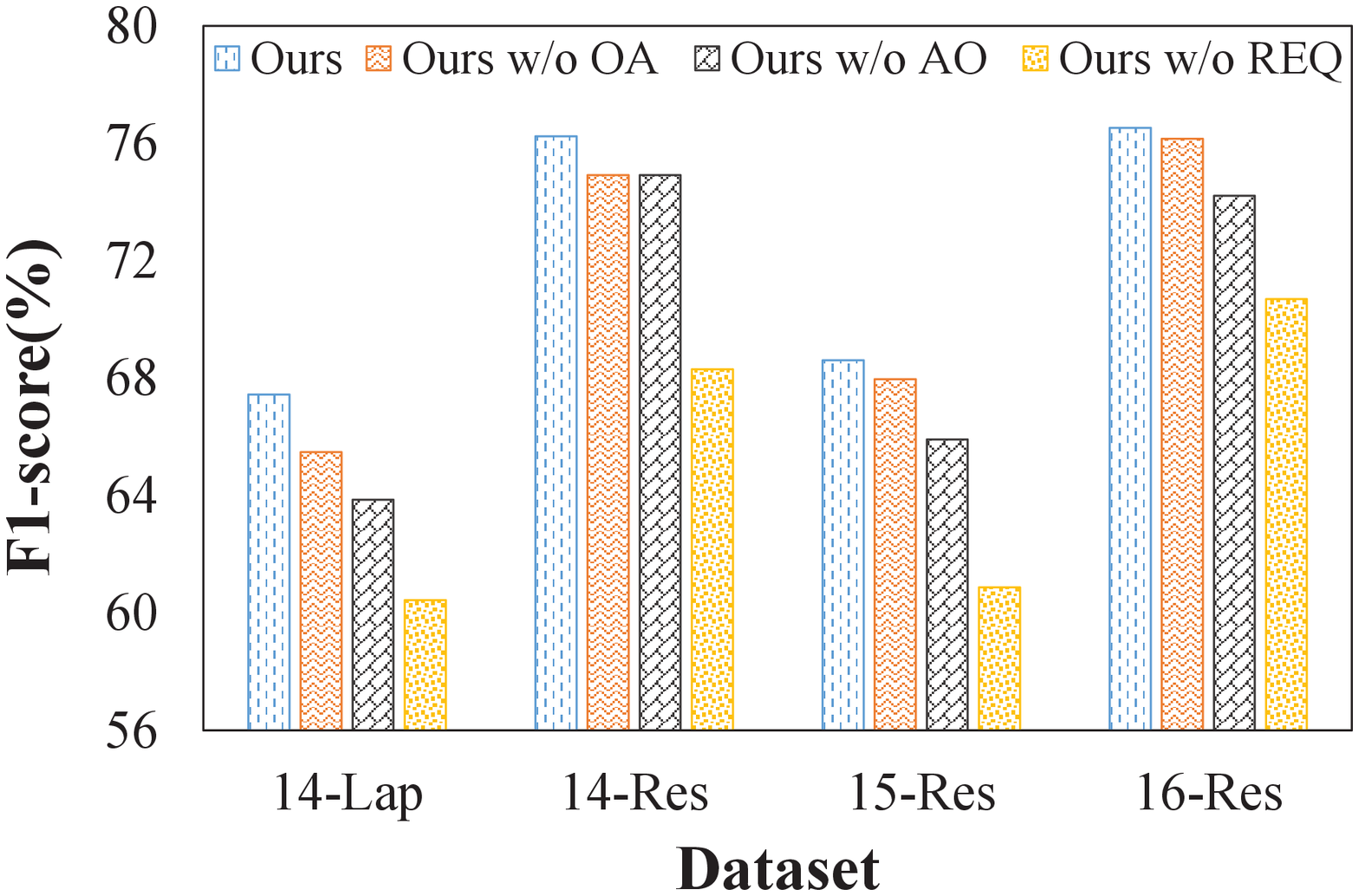}
(c) Aspect-Opinion Pair Extraction\\
\end{minipage}
\begin{minipage}{0.35\textwidth}
\centering
\includegraphics[width=1\textwidth]{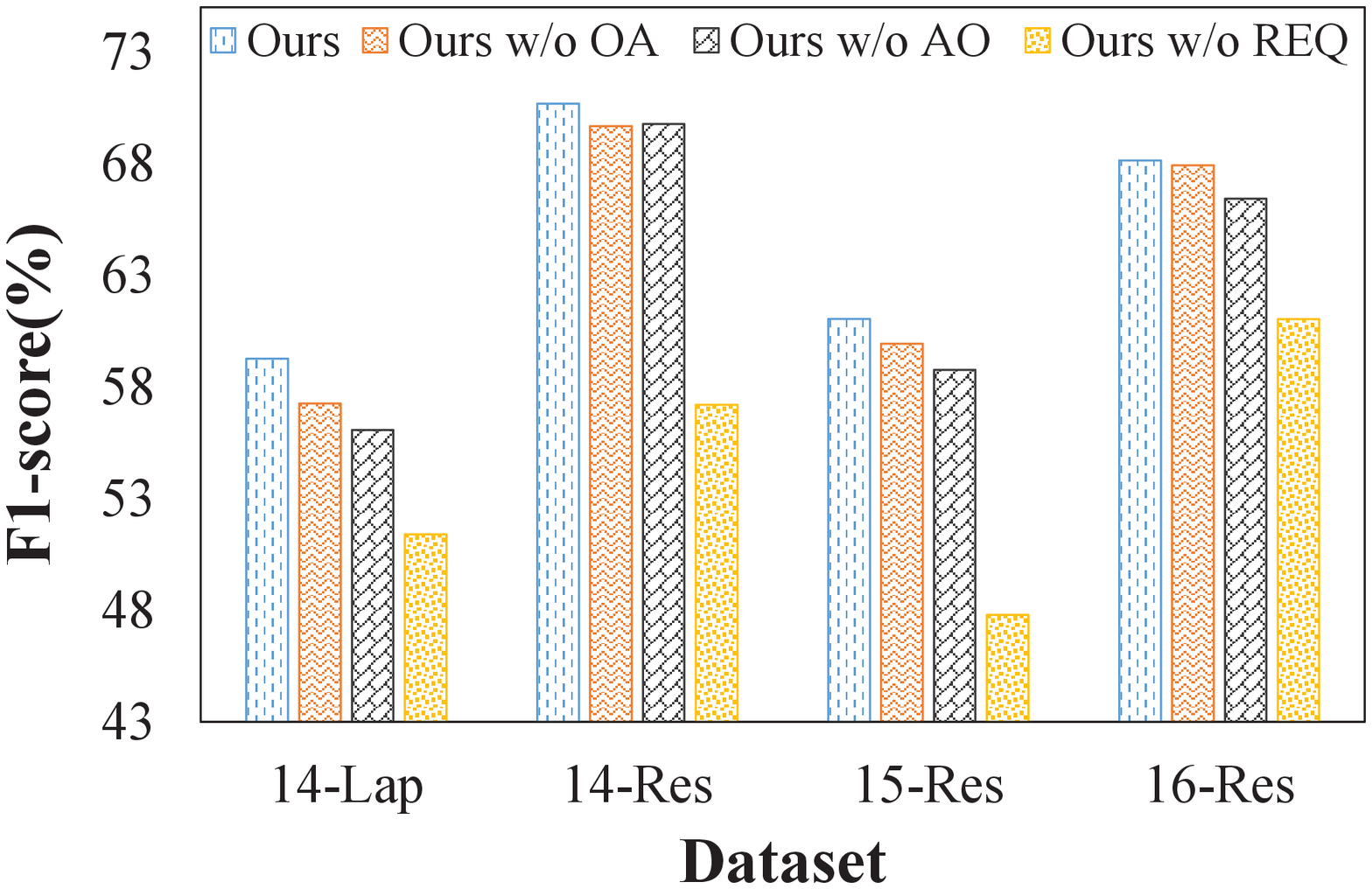}
(d) Triplet Extraction\\
\end{minipage}
\caption{Experimental results of ablation study on the restrictive extraction query and the bidirectional structure.}\label{Thred}
\end{figure*}

We first validate whether the restrictive extraction query could effectively capture and exploit the dependency between opinion entity extraction and relation detection for better performance. Accordingly, we construct a two-stage model similar to TSF, called `Ours w/o REQ'. In the first stage, we remove the restrictive extraction query $Q^{\mathcal{R}}$ from BMRC for only the opinion entity extraction and sentiment classification. The stage-2 model, which is responsible for relation detection, is also based on MRC with the input query `\textit{Matched 
the aspect $a_{i}$ and the opinion expression $o_{j}$?}'. Experimental results are shown in Figure \ref{Thred}. Although the performances on aspect extraction and opinion extraction are comparable, the performances of `Ours w/o REQ' on triplet extraction and aspect-opinion pair extraction are evidently inferior than BMRC. The reason is that with the removal of the restrictive extraction query, the opinion entity extraction and relation detection are separated and no dependency would be captured by `Ours w/o REQ'.
This indicates the effectiveness of the restrictive query at capturing the dependency. 
\begin{table}[]
\centering
\scalebox{0.95}{
\begin{tabular}{l|c|c|c|c}
\hline
   Models               & 14-Lap*       & 14-Res*     & 15-Res*           & 16-Res*          \\ \hline
IMN + IOG   & 47.68      &  61.65         & 53.75         &   -            \\ 
GTS   & 54.58      &  \textbf{70.20}          & 58.67         &    \textbf{67.58}            \\ 
Ours            & \textbf{57.83}    &  70.01       &  \textbf{58.74}   & 67.49              \\ \hline 
\end{tabular}
}
\caption{Experimental results of aspect sentiment triplet extraction on the AFOE datasets. (\textit{F$_{1}$-score, \%}). }\label{AFOE}
\end{table}

\begin{table}[]
\centering
\scalebox{0.9}{
\begin{tabular}{c|c|c|c|c}
\hline
\multirow{2}{*}{Datasets} & \multicolumn{2}{c|}{A} & \multicolumn{2}{c}{A-S} \\ \cline{2-5} 
                  & Ours       & Our w/o REQ     & Ours           & Ours w/o REQ          \\ \hline
14-Lap            & 78.94      & \textbf{80.06}          & \textbf{67.27}         & 61.61                \\ 
14-Res            & \textbf{83.31}      & 82.73          & \textbf{76.39}         & 66.26                \\ 
15-Res            & 75.67      & \textbf{79.00}          & \textbf{67.16}      & 56.82                \\ 
16-Res            & \textbf{83.28}      & 80.60           & \textbf{73.18}         & 68.82                \\ \hline
\end{tabular}
}
\caption{Experimental results of the ablation study on relation-aware sentiment classification (\textit{F$_{1}$-score, \%}). Specifically, `A' and `A-S' stand for aspect term extraction and aspect term and sentiment co-extraction, respectively. }\label{SA}
\end{table}
\subsection{Effect of the Bidirectional MRC Structure}
To explore the effect of bidirectional MRC structure, we compare our model with two unidirectional models, including `Ours w/o AO' and `Ours w/o OA'. Concretely, `Ours w/o AO' extracts triplets only through O$\rightarrow$A direction, and `Ours w/o OA' extracts triplets through A$\rightarrow$O direction. As shown in Figure \ref{Thred}, `Ours w/o OA' shows inferior performance on opinion term extraction without O$\rightarrow$A direction MRC, while `Ours w/o AO' shows worse performance on aspect term extraction. This further harms the performances on aspect-opinion pair extraction and triplet extraction. 
The reason is that both aspects and opinion expressions can initiate aspect-opinion pairs, and the model will be biased when relations are forced to be detected by either aspects or opinions only. 
By introducing the bidirectional design, the two direction MRCs can complement each other and further improve the performance of aspect-opinion pair extraction and triplet extraction.

\subsection{Effect of Relation-Aware Sentiment Classification}
In order to examine the benefit that the relations between aspects and opinion expressions provide for the sentiment classification, we compare the performances of our model and `Ours w/o REQ'. Experimental results on aspect term extraction and aspect term and sentiment co-extraction are shown in Table \ref{SA}. Since `Ours w/o REQ' separate relation detection and sentiment classification in two stages, the detected relations cannot directly provide assistance to sentiment classification.
According to the results, although removing relation detection from joint learning does not harm the performance of aspect term extraction seriously, the performances of aspect term and sentiment co-extraction are all significantly weakened. This clearly indicates that the relations between aspects and opinion expressions can effectively boost the performance of sentiment classification.

\begin{table}[]
\centering
\begin{tabular}{l|c|c|c|c}
\hline
   Models               & 14-Lap       & 14-Res     & 15-Res           & 16-Res          \\ \hline
TSF            & 43.50      &  51.89         &  46.79        &   53.62              \\ 
Ours w/o BERT   & 48.15      &  63.32         & 53.77         &   63.16            \\ 
Ours w/o REQ   & 51.40      &  57.20         & 47.79         &   61.03            \\ 
Ours            & \textbf{59.27}    &  \textbf{70.69}        &  \textbf{61.05}   & \textbf{68.13}               \\ \hline 
\end{tabular}
\caption{Experimental results of the ablation study on aspect sentiment triplet extraction  (\textit{F$_{1}$-score, \%}), which aims to analyze the effect of BERT. }\label{BERT}
\end{table}

\subsection{Effect of BERT}
We analyze the effect of BERT and our contributions from two perspectives. First, we construct our model based on BiDAF, which is a typical reading comprehension model without BERT, and refer it to `Ours w/o BERT'. According to the results shown in Table \ref{BERT}, it significantly surpasses TSF by an average of 8.15\% F1-score on triplet extraction, which shows that our model can achieve SOTA performance without BERT. Besides, compared with `Ours w/o BERT', our model further improves 7.69\% F1-score, which is brought by BERT.
Second, we compare our model with 'Ours w/o REQ' and TSF. The ablation model 'Ours w/o REQ' can be regarded as an implementation version of TSF based on BERT and MRC framework. By comparing it with TSF, the results show that BERT based MRC model can bring an average of 5.41\% F1-score improvement on triplet extraction against the counterpart model without BERT. By further introducing the bidirectional MRC structure and three types of queries, our model further outperforms 'Ours w/o REQ' by 10.4\% F1-score. These two-fold analyses indicate that our contributions play a greater role in improving performances than BERT.


\section{Conclusion}

In this paper, we formalized the aspect sentiment triplet extraction (ASTE) task as a multi-turn machine reading comprehension (MTMRC) task and proposed the bidirectional MRC (BMRC) framework with well-designed queries. Specifically, the non-restrictive and restrictive extraction queries are designed to naturally fuse opinion entity extraction and relation detection, enhancing the dependency between them. By devising the bidirectional MRC structure, it can be ensured that either an aspect or an opinion expression can trigger an aspect-opinion pair just like human's reading behavior. In addition, the sentiment classification query and joint learning manner 
are used to further promote sentiment classification with the incorporation of relations between aspects and opinion expressions. The empirical study demonstrated that our model achieves state-of-the-art performance.

\section{Acknowledgments}
This research is supported by the National Natural Science Foundation of China under grant No. 61976119 and the Major Program of Science and Technology of Tianjin under grant No. 18ZXZNGX00310.

\bibliographystyle{aaai21}
\bibliography{aaai2021_ref}

\begin{thebibliography}{47}
\providecommand{\natexlab}[1]{#1}
\providecommand{\url}[1]{\texttt{#1}}
\providecommand{\urlprefix}{URL }
\expandafter\ifx\csname urlstyle\endcsname\relax
  \providecommand{\doi}[1]{doi:\discretionary{}{}{}#1}\else
  \providecommand{\doi}{doi:\discretionary{}{}{}\begingroup
  \urlstyle{rm}\Url}\fi

\bibitem[{Chen et~al.(2020)Chen, Liu, Wang, Zhang, and Chi}]{SDRN}
Chen, S.; Liu, J.; Wang, Y.; Zhang, W.; and Chi, Z. 2020.
\newblock Synchronous Double-channel Recurrent Network for Aspect-Opinion Pair
  Extraction.
\newblock In \emph{ACL 2020}, 6515--6524.

\bibitem[{Chen and Qian(2020)}]{RACL2020}
Chen, Z.; and Qian, T. 2020.
\newblock Relation-Aware Collaborative Learning for Unified Aspect-Based
  Sentiment Analysis.
\newblock In \emph{ACL 2020}, 3685--3694.

\bibitem[{Dai and Song(2019)}]{DBLP:conf/acl/DaiS19}
Dai, H.; and Song, Y. 2019.
\newblock Neural Aspect and Opinion Term Extraction with Mined Rules as Weak
  Supervision.
\newblock In \emph{ACL 2019}, 5268--5277.

\bibitem[{Devlin et~al.(2019)Devlin, Chang, Lee, and
  Toutanova}]{DBLP:conf/naacl/DevlinCLT19}
Devlin, J.; Chang, M.; Lee, K.; and Toutanova, K. 2019.
\newblock {BERT:} Pre-training of Deep Bidirectional Transformers for Language
  Understanding.
\newblock In \emph{NAACL-HLT 2019}, 4171--4186.

\bibitem[{Dong et~al.(2014)Dong, Wei, Tan, Tang, Zhou, and
  Xu}]{DBLP:conf/acl/DongWTTZX14}
Dong, L.; Wei, F.; Tan, C.; Tang, D.; Zhou, M.; and Xu, K. 2014.
\newblock Adaptive Recursive Neural Network for Target-dependent Twitter
  Sentiment Classification.
\newblock In \emph{ACL 2014}, 49--54.

\bibitem[{Fan et~al.(2019)Fan, Wu, Dai, Huang, and
  Chen}]{DBLP:conf/naacl/FanWDHC19}
Fan, Z.; Wu, Z.; Dai, X.; Huang, S.; and Chen, J. 2019.
\newblock Target-oriented Opinion Words Extraction with Target-fused Neural
  Sequence Labeling.
\newblock In \emph{NAACL-HLT 2019}, 2509--2518.

\bibitem[{Hazarika et~al.(2018)Hazarika, Poria, Vij, Krishnamurthy, Cambria,
  and Zimmermann}]{DBLP:conf/naacl/HazarikaPVKCZ18}
Hazarika, D.; Poria, S.; Vij, P.; Krishnamurthy, G.; Cambria, E.; and
  Zimmermann, R. 2018.
\newblock Modeling Inter-Aspect Dependencies for Aspect-Based Sentiment
  Analysis.
\newblock In \emph{NAACL 2018}, 266--270.

\bibitem[{He et~al.(2017)He, Lee, Ng, and Dahlmeier}]{DBLP:conf/acl/HeLND17}
He, R.; Lee, W.~S.; Ng, H.~T.; and Dahlmeier, D. 2017.
\newblock An Unsupervised Neural Attention Model for Aspect Extraction.
\newblock In \emph{ACL 2017}, 388--397.

\bibitem[{He et~al.(2018)He, Lee, Ng, and Dahlmeier}]{DBLP:conf/acl/HeLND18}
He, R.; Lee, W.~S.; Ng, H.~T.; and Dahlmeier, D. 2018.
\newblock Exploiting Document Knowledge for Aspect-level Sentiment
  Classification.
\newblock In \emph{ACL 2018}, 579--585.

\bibitem[{He et~al.(2019)He, Lee, Ng, and Dahlmeier}]{he2019interactive}
He, R.; Lee, W.~S.; Ng, H.~T.; and Dahlmeier, D. 2019.
\newblock An Interactive Multi-Task Learning Network for End-to-End
  Aspect-Based Sentiment Analysis.
\newblock In \emph{ACL 2019}, 504--515.

\bibitem[{Hu et~al.(2019)Hu, Zhao, Zhang, Cai, Su, Cheng, and
  Shen}]{DBLP:conf/emnlp/HuZZCSCS19}
Hu, M.; Zhao, S.; Zhang, L.; Cai, K.; Su, Z.; Cheng, R.; and Shen, X. 2019.
\newblock {CAN:} Constrained Attention Networks for Multi-Aspect Sentiment
  Analysis.
\newblock In \emph{EMNLP-IJCNLP 2019}, 4600--4609.

\bibitem[{Levy et~al.(2017)Levy, Seo, Choi, and
  Zettlemoyer}]{DBLP:conf/conll/LevySCZ17}
Levy, O.; Seo, M.; Choi, E.; and Zettlemoyer, L. 2017.
\newblock Zero-Shot Relation Extraction via Reading Comprehension.
\newblock In \emph{CoNLL 2017}, 333--342.

\bibitem[{Li and Lu(2017)}]{DBLP:conf/aaai/LiL17}
Li, H.; and Lu, W. 2017.
\newblock Learning Latent Sentiment Scopes for Entity-Level Sentiment Analysis.
\newblock In \emph{AAAI 2017}, 3482--3489.

\bibitem[{Li et~al.(2019{\natexlab{a}})Li, Bing, Li, and Lam}]{LiBLL19}
Li, X.; Bing, L.; Li, P.; and Lam, W. 2019{\natexlab{a}}.
\newblock A Unified Model for Opinion Target Extraction and Target Sentiment
  Prediction.
\newblock In \emph{AAAI 2019}, 6714--6721.

\bibitem[{Li et~al.(2018)Li, Bing, Li, Lam, and Yang}]{Li2018}
Li, X.; Bing, L.; Li, P.; Lam, W.; and Yang, Z. 2018.
\newblock Aspect Term Extraction with History Attention and Selective
  Transformation.
\newblock In \emph{IJCAI 2018}, 4194--4200.

\bibitem[{Li et~al.(2020)Li, Feng, Meng, Han, Wu, and Li}]{li2019unified}
Li, X.; Feng, J.; Meng, Y.; Han, Q.; Wu, F.; and Li, J. 2020.
\newblock A Unified {MRC} Framework for Named Entity Recognition.
\newblock In \emph{ACL 2020}, 5849--5859.

\bibitem[{Li and Lam(2017)}]{DBLP:conf/emnlp/LiL17}
Li, X.; and Lam, W. 2017.
\newblock Deep Multi-Task Learning for Aspect Term Extraction with Memory
  Interaction.
\newblock In \emph{EMNLP 2017}, 2886--2892.

\bibitem[{Li et~al.(2019{\natexlab{b}})Li, Yin, Sun, Li, Yuan, Chai, Zhou, and
  Li}]{DBLP:conf/acl/LiYSLYCZL19}
Li, X.; Yin, F.; Sun, Z.; Li, X.; Yuan, A.; Chai, D.; Zhou, M.; and Li, J.
  2019{\natexlab{b}}.
\newblock Entity-Relation Extraction as Multi-Turn Question Answering.
\newblock In \emph{ACL 2019}, 1340--1350.

\bibitem[{Li et~al.(2019{\natexlab{c}})Li, Wei, Zhang, Zhang, and
  Li}]{DBLP:conf/aaai/LiW0Z019}
Li, Z.; Wei, Y.; Zhang, Y.; Zhang, X.; and Li, X. 2019{\natexlab{c}}.
\newblock Exploiting Coarse-to-Fine Task Transfer for Aspect-Level Sentiment
  Classification.
\newblock In \emph{AAAI 2019}, 4253--4260.

\bibitem[{Liu, Xu, and Zhao(2012)}]{DBLP:conf/emnlp/LiuXZ12}
Liu, K.; Xu, L.; and Zhao, J. 2012.
\newblock Opinion Target Extraction Using Word-Based Translation Model.
\newblock In Tsujii, J.; Henderson, J.; and Pasca, M., eds., \emph{EMNLP-CoNLL
  2012}, 1346--1356.

\bibitem[{Liu, Xu, and Zhao(2015)}]{DBLP:journals/tkde/LiuXZ15}
Liu, K.; Xu, L.; and Zhao, J. 2015.
\newblock Co-Extracting Opinion Targets and Opinion Words from Online Reviews
  Based on the Word Alignment Model.
\newblock \emph{{IEEE} Trans. Knowl. Data Eng.} 27(3): 636--650.

\bibitem[{Liu, Joty, and Meng(2015)}]{liu2015fine}
Liu, P.; Joty, S.; and Meng, H. 2015.
\newblock Fine-grained opinion mining with recurrent neural networks and word
  embeddings.
\newblock In \emph{EMNLP 2015}, 1433--1443.

\bibitem[{Loshchilov and Hutter(2017)}]{DBLP:journals/corr/abs-1711-05101}
Loshchilov, I.; and Hutter, F. 2017.
\newblock Fixing Weight Decay Regularization in Adam.
\newblock \emph{CoRR} abs/1711.05101.

\bibitem[{Luo et~al.(2019)Luo, Li, Liu, and Zhang}]{LuoLLZ19}
Luo, H.; Li, T.; Liu, B.; and Zhang, J. 2019.
\newblock {DOER:} Dual Cross-Shared {RNN} for Aspect Term-Polarity
  Co-Extraction.
\newblock In \emph{ACL 2019}, 591--601.

\bibitem[{Ma et~al.(2019)Ma, Li, Wu, Xie, and Wang}]{MaLWXW19}
Ma, D.; Li, S.; Wu, F.; Xie, X.; and Wang, H. 2019.
\newblock Exploring Sequence-to-Sequence Learning in Aspect Term Extraction.
\newblock In \emph{ACL 2019}, 3538--3547.

\bibitem[{Ma et~al.(2017)Ma, Li, Zhang, and Wang}]{DBLP:conf/ijcai/MaLZW17}
Ma, D.; Li, S.; Zhang, X.; and Wang, H. 2017.
\newblock Interactive Attention Networks for Aspect-Level Sentiment
  Classification.
\newblock In \emph{IJCAI 2017}, 4068--4074.

\bibitem[{McCann et~al.(2018)McCann, Keskar, Xiong, and
  Socher}]{DBLP:journals/corr/abs-1806-08730}
McCann, B.; Keskar, N.~S.; Xiong, C.; and Socher, R. 2018.
\newblock The Natural Language Decathlon: Multitask Learning as Question
  Answering.
\newblock \emph{CoRR} abs/1806.08730.

\bibitem[{Mitchell et~al.(2013)Mitchell, Aguilar, Wilson, and
  Durme}]{DBLP:conf/emnlp/MitchellAWD13}
Mitchell, M.; Aguilar, J.; Wilson, T.; and Durme, B.~V. 2013.
\newblock Open Domain Targeted Sentiment.
\newblock In \emph{EMNLP 2013}, 1643--1654.

\bibitem[{Nguyen and Shirai(2015)}]{DBLP:conf/emnlp/NguyenS15}
Nguyen, T.~H.; and Shirai, K. 2015.
\newblock PhraseRNN: Phrase Recursive Neural Network for Aspect-based Sentiment
  Analysis.
\newblock In \emph{EMNLP 2015}, 2509--2514.

\bibitem[{Peng et~al.(2020)Peng, Xu, Bing, Huang, Lu, and
  Si}]{DBLP:conf/aaai/PengXBHLS20}
Peng, H.; Xu, L.; Bing, L.; Huang, F.; Lu, W.; and Si, L. 2020.
\newblock Knowing What, How and Why: {A} Near Complete Solution for
  Aspect-Based Sentiment Analysis.
\newblock In \emph{AAAI 2020}, 8600--8607.

\bibitem[{Peters et~al.(2018)Peters, Neumann, Iyyer, Gardner, Clark, Lee, and
  Zettlemoyer}]{DBLP:conf/naacl/PetersNIGCLZ18}
Peters, M.~E.; Neumann, M.; Iyyer, M.; Gardner, M.; Clark, C.; Lee, K.; and
  Zettlemoyer, L. 2018.
\newblock Deep Contextualized Word Representations.
\newblock In \emph{NAACL-HLT 2018}, 2227--2237.

\bibitem[{Pontiki et~al.(2016)Pontiki, Galanis, Papageorgiou, Androutsopoulos,
  Manandhar, Al{-}Smadi, Al{-}Ayyoub, Zhao, Qin, Clercq, Hoste, Apidianaki,
  Tannier, Loukachevitch, Kotelnikov, Bel, Zafra, and
  Eryigit}]{DBLP:conf/semeval/PontikiGPAMAAZQ16}
Pontiki, M.; Galanis, D.; Papageorgiou, H.; Androutsopoulos, I.; Manandhar, S.;
  Al{-}Smadi, M.; Al{-}Ayyoub, M.; Zhao, Y.; Qin, B.; Clercq, O.~D.; Hoste, V.;
  Apidianaki, M.; Tannier, X.; Loukachevitch, N.~V.; Kotelnikov, E.~V.; Bel,
  N.; Zafra, S. M.~J.; and Eryigit, G. 2016.
\newblock SemEval-2016 Task 5: Aspect Based Sentiment Analysis.
\newblock In \emph{Proceedings of the 10th International Workshop on Semantic
  Evaluation, SemEval@NAACL-HLT 2016}, 19--30.

\bibitem[{Pontiki et~al.(2015)Pontiki, Galanis, Papageorgiou, Manandhar, and
  Androutsopoulos}]{DBLP:conf/semeval/PontikiGPMA15}
Pontiki, M.; Galanis, D.; Papageorgiou, H.; Manandhar, S.; and Androutsopoulos,
  I. 2015.
\newblock SemEval-2015 Task 12: Aspect Based Sentiment Analysis.
\newblock In \emph{Proceedings of the 9th International Workshop on Semantic
  Evaluation, SemEval@NAACL-HLT 2015}, 486--495.

\bibitem[{Pontiki et~al.(2014)Pontiki, Galanis, Pavlopoulos, Papageorgiou,
  Androutsopoulos, and Manandhar}]{DBLP:conf/semeval/PontikiGPPAM14}
Pontiki, M.; Galanis, D.; Pavlopoulos, J.; Papageorgiou, H.; Androutsopoulos,
  I.; and Manandhar, S. 2014.
\newblock SemEval-2014 Task 4: Aspect Based Sentiment Analysis.
\newblock In \emph{Proceedings of the 8th International Workshop on Semantic
  Evaluation, SemEval@COLING 2014}, 27--35.

\bibitem[{Poria, Cambria, and Gelbukh(2016)}]{poria2016aspect}
Poria, S.; Cambria, E.; and Gelbukh, A. 2016.
\newblock Aspect extraction for opinion mining with a deep convolutional neural
  network.
\newblock \emph{Knowledge-Based Systems} 108: 42--49.

\bibitem[{Radford et~al.(2019)Radford, Wu, Child, Luan, Amodei, and
  Sutskever}]{radford2019language}
Radford, A.; Wu, J.; Child, R.; Luan, D.; Amodei, D.; and Sutskever, I. 2019.
\newblock Language models are unsupervised multitask learners.
\newblock \emph{OpenAI Blog 2019} 1(8): 9.

\bibitem[{Seo et~al.(2017)Seo, Kembhavi, Farhadi, and
  Hajishirzi}]{DBLP:conf/iclr/SeoKFH17}
Seo, M.~J.; Kembhavi, A.; Farhadi, A.; and Hajishirzi, H. 2017.
\newblock Bidirectional Attention Flow for Machine Comprehension.
\newblock In \emph{ICLR 2017}.

\bibitem[{Sun et~al.(2019)Sun, Zhang, Mensah, Mao, and
  Liu}]{DBLP:conf/emnlp/SunZMML19}
Sun, K.; Zhang, R.; Mensah, S.; Mao, Y.; and Liu, X. 2019.
\newblock Aspect-Level Sentiment Analysis Via Convolution over Dependency Tree.
\newblock In \emph{EMNLP-IJCNLP 2019}, 5678--5687.

\bibitem[{Tang, Qin, and Liu(2016)}]{DBLP:conf/emnlp/TangQL16}
Tang, D.; Qin, B.; and Liu, T. 2016.
\newblock Aspect Level Sentiment Classification with Deep Memory Network.
\newblock In \emph{EMNLP 2016}, 214--224.

\bibitem[{Wang et~al.(2018)Wang, Mazumder, Liu, Zhou, and
  Chang}]{DBLP:conf/acl/LiuCWMZ18}
Wang, S.; Mazumder, S.; Liu, B.; Zhou, M.; and Chang, Y. 2018.
\newblock Target-Sensitive Memory Networks for Aspect Sentiment Classification.
\newblock In \emph{ACL 2018}, 957--967.

\bibitem[{Wang et~al.(2016)Wang, Pan, Dahlmeier, and Xiao}]{Wang2016}
Wang, W.; Pan, S.~J.; Dahlmeier, D.; and Xiao, X. 2016.
\newblock Recursive Neural Conditional Random Fields for Aspect-based Sentiment
  Analysis.
\newblock In \emph{EMNLP 2016}, 616--626.

\bibitem[{Wang et~al.(2017)Wang, Pan, Dahlmeier, and
  Xiao}]{DBLP:conf/aaai/WangPDX17}
Wang, W.; Pan, S.~J.; Dahlmeier, D.; and Xiao, X. 2017.
\newblock Coupled Multi-Layer Attentions for Co-Extraction of Aspect and
  Opinion Terms.
\newblock In \emph{AAAI 2017}, 3316--3322.

\bibitem[{Wu et~al.(2020{\natexlab{a}})Wu, Ying, Zhao, Fan, Dai, and
  Xia}]{Wu2020EMNLP}
Wu, Z.; Ying, C.; Zhao, F.; Fan, Z.; Dai, X.; and Xia, R. 2020{\natexlab{a}}.
\newblock Grid Tagging Scheme for Aspect-oriented Fine-grained Opinion
  Extraction.
\newblock In \emph{Findings of the Association for Computational Linguistics:
  EMNLP 2020}, 2576--2585.

\bibitem[{Wu et~al.(2020{\natexlab{b}})Wu, Zhao, Dai, Huang, and Chen}]{Wu2020}
Wu, Z.; Zhao, F.; Dai, X.; Huang, S.; and Chen, J. 2020{\natexlab{b}}.
\newblock Latent Opinions Transfer Network for Target-Oriented Opinion Words
  Extraction.
\newblock In \emph{AAAI 2020}, 9298--9305.

\bibitem[{Xu et~al.(2018)Xu, Liu, Shu, and Yu}]{DECNN}
Xu, H.; Liu, B.; Shu, L.; and Yu, P.~S. 2018.
\newblock Double Embeddings and CNN-based Sequence Labeling for Aspect
  Extraction.
\newblock In \emph{ACL 2018}, 592--598.

\bibitem[{Yu et~al.(2018)Yu, Dohan, Luong, Zhao, Chen, Norouzi, and
  Le}]{DBLP:conf/iclr/YuDLZ00L18}
Yu, A.~W.; Dohan, D.; Luong, M.; Zhao, R.; Chen, K.; Norouzi, M.; and Le, Q.~V.
  2018.
\newblock QANet: Combining Local Convolution with Global Self-Attention for
  Reading Comprehension.
\newblock In \emph{ICLR 2018}.

\bibitem[{Zhao et~al.(2020)Zhao, Huang, Zhang, Lu, and Xue}]{SpanMIT}
Zhao, H.; Huang, L.; Zhang, R.; Lu, Q.; and Xue, H. 2020.
\newblock SpanMlt: {A} Span-based Multi-Task Learning Framework for Pair-wise
  Aspect and Opinion Terms Extraction.
\newblock In \emph{ACL 2020}, 3239--3248.

\end{thebibliography}

\end{document}